%% file: root.tex
\documentclass[letterpaper, 10 pt, conference]{ieeeconf}  % Comment this line out if you need a4paper

\IEEEoverridecommandlockouts                              % This command is only needed if 
\usepackage{epsfig}
\usepackage{multirow}
\usepackage{makecell}
\usepackage{amsmath}
\usepackage{amssymb}
\usepackage{times}
\usepackage{pifont}

\usepackage[ruled,vlined]{algorithm2e}

\usepackage{xcolor}
\usepackage{esvect}

\title{\LARGE \bf
PanoNet: Real-time Panoptic Segmentation through Position-Sensitive Feature Embedding
}

\author{Xia Chen$^{1}$, Jianren Wang$^{1}$, Martial Hebert$^{1}$% <-this % stops a space
\thanks{$^{1}$Xia Chen, Jianren Wang, Martial Hebert are with the Robotics Institute, Carnegie Mellon University, 5000 Forbes Ave., Pittsburgh, PA 15213, USA
        {\tt\small xiac, jianrenw, mhebert@andrew.cmu.edu}}%
}

\begin{document}

\maketitle
\thispagestyle{empty}
\pagestyle{empty}

%%%%%%%%%%%%%%%%%%%%%%%%%%%%%%%%%%%%%%%%%%%%%%%%%%%%%%%%%%%%%%%%%%%%%%%%%%%%%%%%
\begin{abstract}

We propose a simple, fast, and flexible framework to generate simultaneously semantic and instance masks  for panoptic segmentation. Our method, called PanoNet, incorporates a clean and natural structure design that tackles the problem purely as a segmentation task without the time-consuming detection process.  We also introduce position-sensitive embedding for instance grouping by accounting for both object's appearance and its spatial location. Overall, PanoNet yields high panoptic quality results of high-resolution Cityscapes images in real-time, significantly faster than all other methods with comparable performance. Our approach well satisfies the practical speed and memory requirement for many applications like autonomous driving and augmented reality.

\end{abstract}

\input{sections/1-introduction.tex}

\input{sections/2-related_work.tex}

\input{sections/3-approach.tex}

\input{sections/4-results.tex}

\input{sections/5-conclusion.tex}

{\small
\bibliographystyle{IEEEtran}
\bibliography{root}
}

\end{document}

%% file: sections/1-introduction.tex
\section{Introduction}

\textit{Things} and \textit{stuff} have long been studied separately: the former formulated as tasks known as object detection or instance segmentation, the latter formulated as tasks known as semantic segmentation. To find an effective design of a unified vision system that generates rich and coherent scene segmentation, panoptic segmentation~\cite{kirillov2018panoptic} was introduced, and it becomes particularly important in autonomous driving and augmented reality~\cite{yao20183d}. 

The uniqueness of panoptic segmentation lies in two aspects: First, this task should be solved efficiently since it needs to be fast in real applications. Second, it unifies the feature presentation and network architecture for semantic segmentation and instance segmentation. However, to our best knowledge, there are no methods satisfying both properties at the same time.
\begin{figure}[h]
  \includegraphics[width=\linewidth]{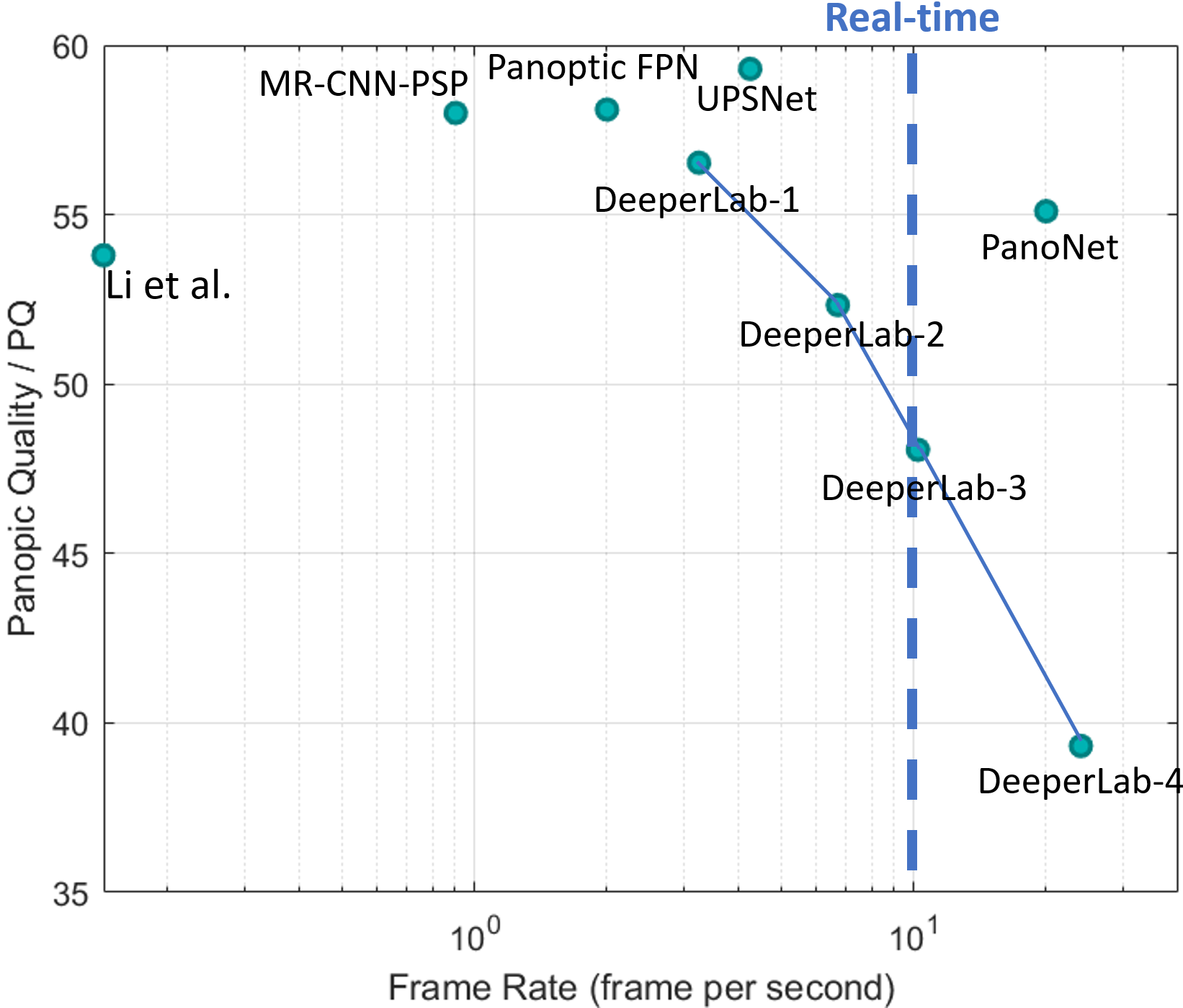}
  \caption{Panoptic quality (PQ) vs inference speed on Cityscapes validation set (full scale 1024 $\times$ 2048 high-resolution image). The listed methods: Li et al.\cite{Li_2018_ECCV}, Mask R-CNN and PSPNet (MR-CNN-PSP)~\cite{kirillov2018panoptic, arxiv:upsnet}, Panoptic FPN~\cite{kirillov2019panoptic}, UPSNet~\cite{arxiv:upsnet}, DeepLab~\cite{yang2019deeperlab} variants and our PanoNet. Our model runs significantly faster than other methods with comparable performance.}
  \label{fig:speed_comparision}
\end{figure}
Although semantic segmentation and instance segmentation are highly relevant, very dissimilar methods were adopted for each task. For semantic segmentation, FCNs with specialized backbones enhanced by dilated convolutions ~\cite{YuKoltun2016,chen2018deeplab} dominate popular leader boards ~\cite{cordts2016cityscapes,everingham2015pascal}. For instance segmentation, region-based Mask R-CNN ~\cite{he2017mask} with a Feature Pyramid Network (FPN) ~\cite{lin2017feature} backbone has been used as a foundation for all top entries in recent recognition challenges ~\cite{Caesar_2018_CVPR, zhou2017scene, neuhold2017mapillary}. To make full use of these top-performing methods, most previous works use two parallel branches, one for instance-level recognition with RPN~\cite{ren2015faster} and one for semantic-level segmentation~\cite{pham2017biseg, kirillov2019panoptic, kirillov2018panoptic}. However, neither inference efficiency and the correlation between these two highly relevant tasks is considered. In these proposed methods, only the feature extraction backbone is shared. Panoptic FPN~\cite{kirillov2019panoptic} predicts instance segmentation masks without using any semantic segmentation result. Similarly, BiSeg~\cite{pham2017biseg} runs R-FCN~\cite{dai2016r} multiple times and applies a max operation to produce the per-pixel likelihood of the object category. 

To alleviate the time inefficiency caused by re-segmentation, our goal is designing a one-stage panoptic segmentation algorithm using FCNs as backbone. Previous work~\cite{de2017semantic} has shown that semantic segmentation frameworks can also be used to distinguish different instances by training the network with a discriminative loss function and then clustering pixel-wise feature embedding into masks of instances. However, it does not take positional information into consideration because of the translational invariance of FCNs. Thus, this approach is unable to segment instances with similar appearance and feature (e.g., two cars of the same model).

We argue that both appearance and position of objects are important when using pixel-wise embedding to cluster instance masks. Inspired by Liu et al.~\cite{liu2018intriguing}, we add spacial information as additional input and train the network to be position-sensitive. Experimental results show that the position-sensitive feature embedding leads to significant improvement of performance. With light-weight backbone like ICNet~\cite{Zhao_2018_ECCV}, we achieve real-time performance and accuracy comparable with the state-of-the-art. Fig. \ref{fig:speed_comparision} shows the speed-accuracy trade-off of panoptic segmentation methods. To the best of our knowledge, our method is the first approach achieving real-time performance on high-resolution images on a single GPU.

% Zamir et al. ~\cite{zamir2018taskonomy} shows high correlation between 2D keypoints detection and semantic segmentation. Thus, out propose PanoNet take both properties of panoptic segmentation into account.

Our key contributions are listed below:
\begin{itemize}
  \item We propose a one-stage panoptic segmentation method (called PanoNet) using highly similar structure and sharable parameters for both semantic and instance segmentation branches. The design of the PanoNet is much simpler than that of other state-of-art solutions.
  \item PanoNet utilizes position of objects on top of their feature embedding for segmentation without region proposals. We develop a unique training procedure that allows the network to learn how the position-sensitive information should be encoded, resulting in significant performance improvement.
  \item PanoNet has excellent speed-accuracy trade-off in terms of both memory and time consumption. With less than 3 GB memory used, PanoNet achieves real-time inference speed (20 FPS) on high-resolution $2048\times1024$ images. On the other hand, PanoNet still has decent performance both visually and on the panoptic quality metrics.
\end{itemize}

% three-fold: First, one-stage. Second, feature embedding. Third, state-of-art.

%% file: sections/2-related_work.tex
\section{Related Works}

%-------------------------------------------------------------------------
\subsection{Semantic Segmentation:}
Semantic segmentation is a classical and well-studied computer vision problem. Convnets have been long used to exploit the contextual information for segmentation~\cite{dai2015convolutional,yao2012describing,havaei2017brain,farabet2013learning,pinheiro2014recurrent,arbelaez2014multiscale}. Recently, a prevalent family of approaches based on Fully Convolutional Networks (FCNs)~\cite{long2015fully} have demonstrated state-of-art performance on several benchmarks~\cite{everingham2015pascal,cordts2016cityscapes,zhou2017scene,Caesar_2018_CVPR}. Four great ideas have been proposed among these methods. The first idea is fusing multi-scale feature~\cite{zhao2017pyramid,chen2016attention,xia2016zoom,hariharan2015hypercolumns}, since higher-layer feature contains more semantic meaning but less local information, and combining multi-scale features can improve the performance. The second idea is using dilated convolution to increase local information and enlarge the receptive field at the same time ~\cite{chen14semantic,chen2018deeplab,chen2017rethinking,chen2018encoder}. The third idea is to adopt probabilistic graphical models (e.g. CRFs) to refine the segmentation result~\cite{chen14semantic,chen2018deeplab,zheng2015conditional,lin2016efficient}. However, this post-processing is time-consuming and breaks the end-to-end modeling. Fourth idea is the usage of Encoder-decoder networks~\cite{badrinarayanan2017segnet,noh2015learning,lin2017refinenet,zhang2018context,bilinski2018dense}. Previous works~\cite{chen2018encoder} show that it can help to obtain sharper segmentations. 

%-------------------------------------------------------------------------
\subsection{Instance Segmentation}
Instance segmentation task is to predict the boundary of each object in the scene. We categorize current instance segmentation methods into two categories: detection-based methods and segmentation-based methods.

For detection-based methods, most models adopt RPN~\cite{ren2015faster} to generate instance proposals. FCIS~\cite{li2017fully} and MaskLab~\cite{Chen_2018_CVPR} utilize the `position-sensitive score map' idea proposed by R-FCN~\cite{dai2016r}. Mask R-CNN~\cite{he2017mask} extends Faster R-CNN by adding a branch for predicting object masks. PANet~\cite{liu2018path} adds a bottom-up path to the FPN backbone aiming at enhancing the localization capability of the entire feature hierarchy and demonstrates outstanding performance.  

The other approach is segmentation-based. The methods that fall into this category always need to learn spatial transformation for instance clustering. Many  methods attempt  to  predict link relationship between each pixel with its neighbors~\cite{shi2017detecting, deng2018pixellink,liang2015proposal} through graphical models~\cite{arnab2017pixelwise,zhang2016instance}, embedding vectors~\cite{newell2017associative} or discriminative loss~\cite{de2017semantic}. InstanceCut~\cite{kirillov2017instancecut} predicts object boundary while Watershed~\cite{bai2017deep} predicts the watershed energy through an end-to-end convolutional neural network and applies multi-cut. Instances are also separated by exploiting depth ordering within an image patch~\cite{zhang2015monocular,uhrig2016pixel}. However, this method requires ground-truth depth maps during training which we do not assume that we have.

Other works focus on fast prediction speed to achieve real-time performance. Box2Pix~\cite{uhrig2018box2pix} relies on a single FCN~\cite{long2015fully} to predict object bounding boxes, pixel-wise semantic object classes and offset vectors toward object centers. However, because of the single architecture and delicate offset vector prediction, the final result is not competitive.

%-------------------------------------------------------------------------
\subsection{Panoptic Segmentation}
\begin{figure*}[ht]
  \includegraphics[width=\linewidth]{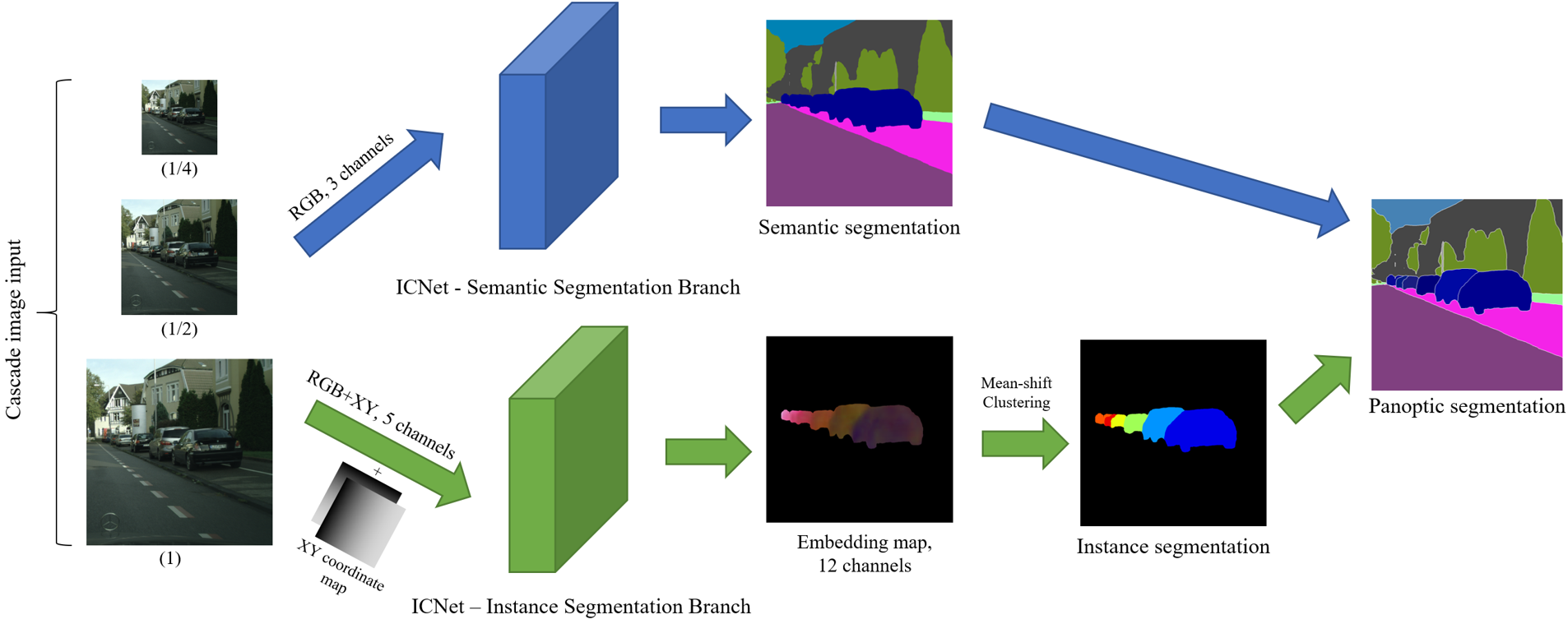}
  \caption{The PanoNet framework for panoptic segmentation. The top branch takes normal images as input and outputs semantic segmentation results. The lower branch takes original images along with additional XY coordinate map and outputs instance segmentation results. The outputs of the two branches can be merged into panoptic segmentation directly without requiring any additional procedures}
  \label{fig:framework}
\end{figure*}
Normally, instance segmentation only focuses several semantic classes (such as humans and cars, usually referred as \textit{things}) and ignores the other (such as road and sky, usually referred as \textit{stuff}). On the other hand, semantic segmentation cannot provide masks of each individual objects. The idea of combining instance and semantic segmentation together is first proposed by Kirillov et al. \cite{kirillov2018panoptic}. The new task, called panoptic segmentation, requires the output to contain both pixel-wise semantic prediction and object boundaries for \textit{things} classes. A simple solution to this problem is proposed in \cite{kirillov2018panoptic} by heuristically combining the instance segmentation results from a Mask R-CNN and semantic segmentation results from a PSPNet. Recently, many works are aimed to find a unified network for the two sub-tasks. Panoptic FPN \cite{kirillov2019panoptic} slightly modifies Mask-RCNN by enabling it to also generate pixel-wise semantic segmentation prediction. UPSNet \cite{arxiv:upsnet} designs a panoptic head with a single network as backbone. However, these methods all rely on a region-proposal based object detector and fail to run in real-time.

In contrast to those methods, we proposed a framework that exploits the strong correlation between detection and segmentation tasks. We use information in both spacial and feature embedding domain for instance grouping to alleviate the inefficiency caused by detection-based segmentation. Our design is simple yet effective. With ICNet~\cite{Zhao_2018_ECCV} as backbone, we achieve $\sim$20-fps inference speed on Cityscapes' full scale images (resolution: $2048\times1024$) with decent prediction accuracy in both semantic and instance segmentation tasks.

%% file: sections/3-approach.tex
\section{PanoNet}
\subsection{Network Structure}

We choose ICNet~\cite{Zhao_2018_ECCV} (originally designed for semantic segmentation) as the backbone of PanoNet. ICNet uses cascade feature fusion unit and cascade label guidance to speed up low-speed segmentation network (PSPNet50) while still maintaining decent accuracy. We modify ICNet by adding an extra branch that takes additional coordinate input channels and outputs an pixel-wise embedding map for instance segmentation. The whole structure of PanoNet is shown in Fig. \ref{fig:framework}.

\begin{figure*}[t]
\centering
        \includegraphics[width=1\columnwidth]{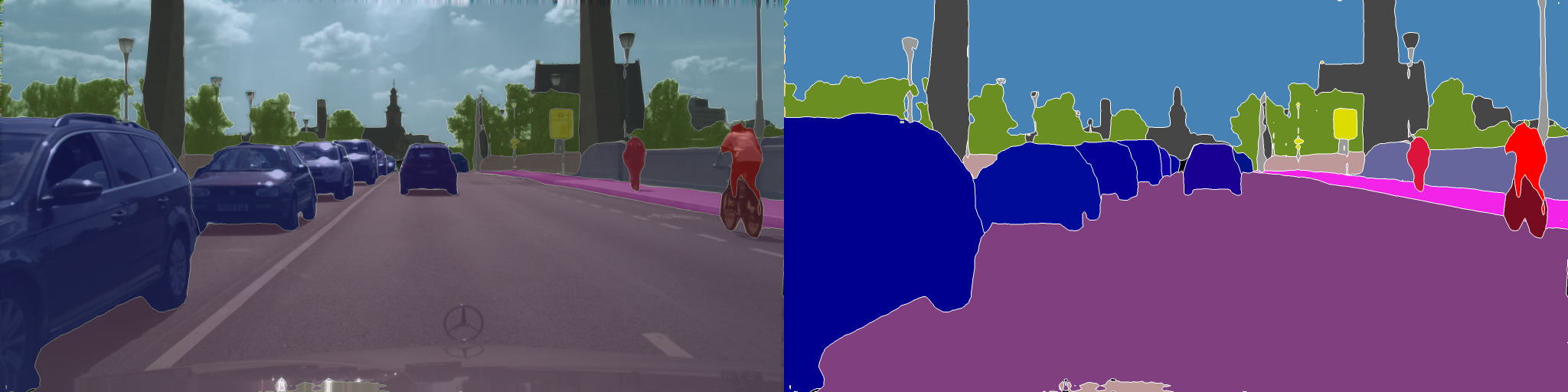}\includegraphics[width=1.0\columnwidth]{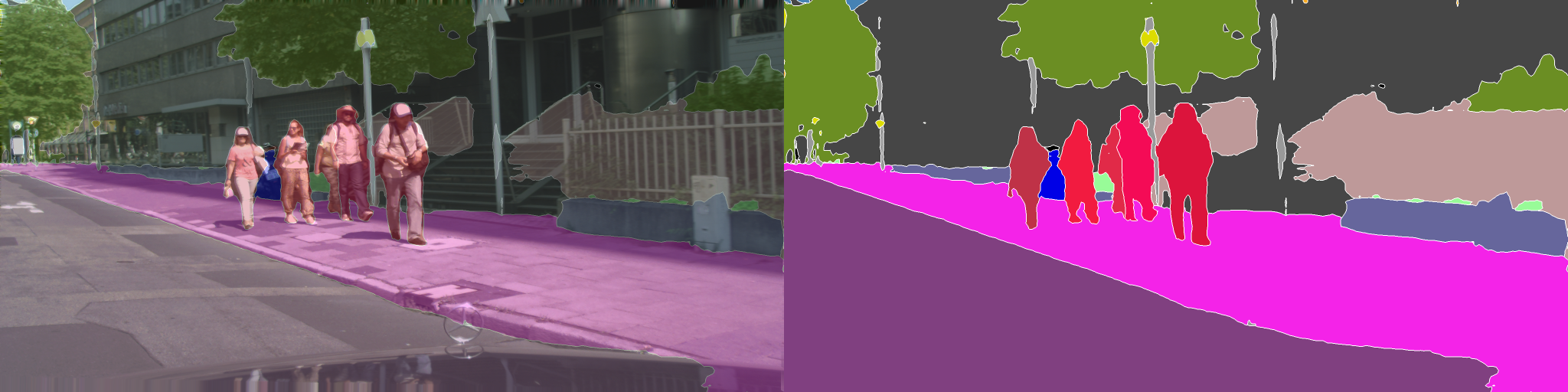}
        \includegraphics[width=1\columnwidth]{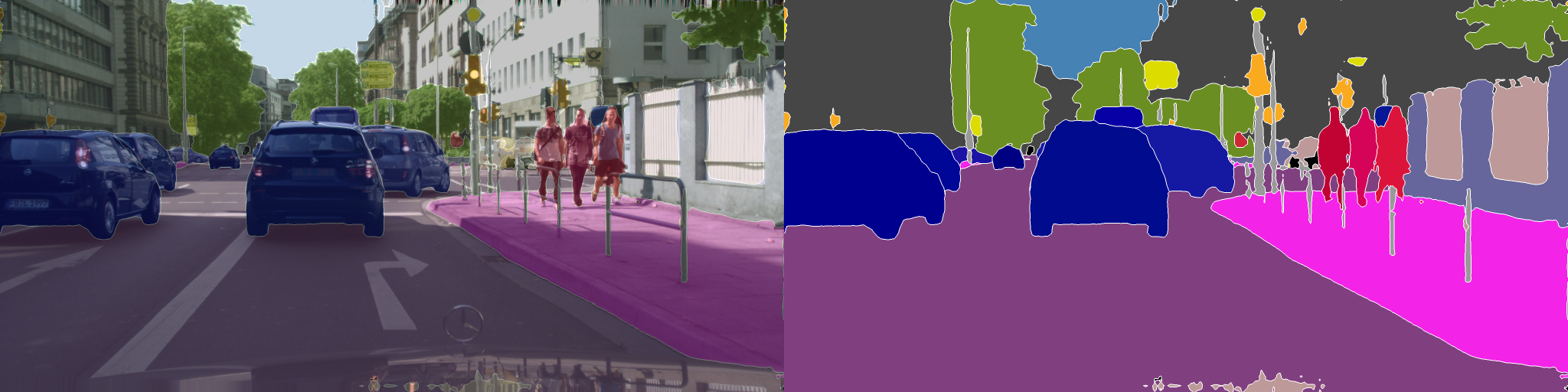}\includegraphics[width=1.0\columnwidth]{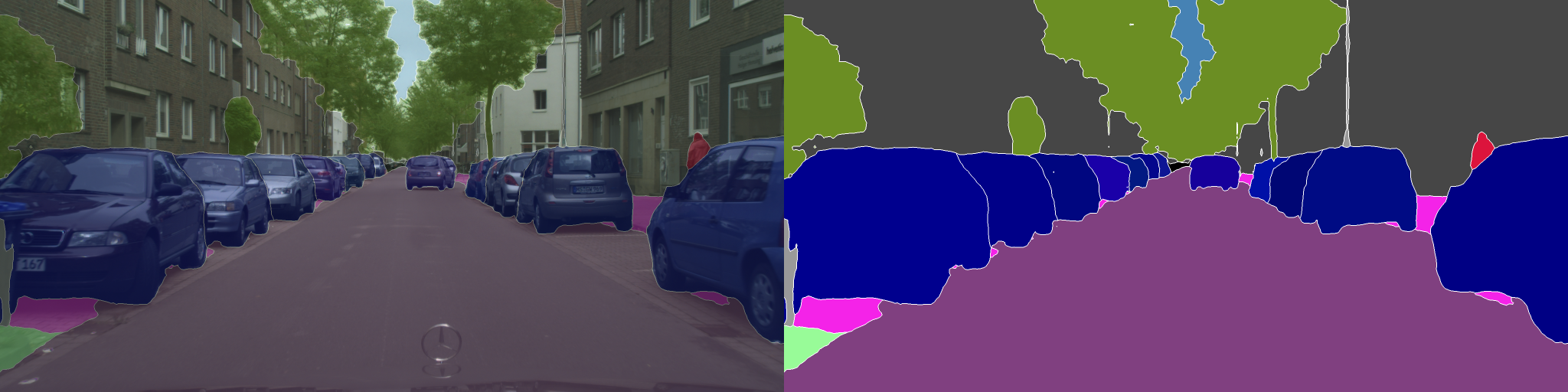}
        \includegraphics[width=1\columnwidth]{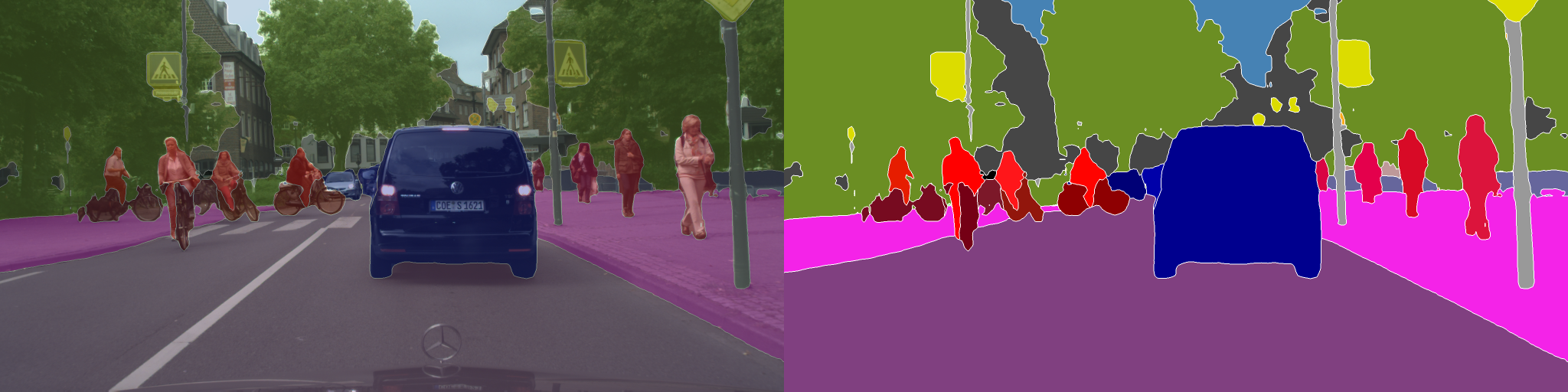}\includegraphics[width=1.0\columnwidth]{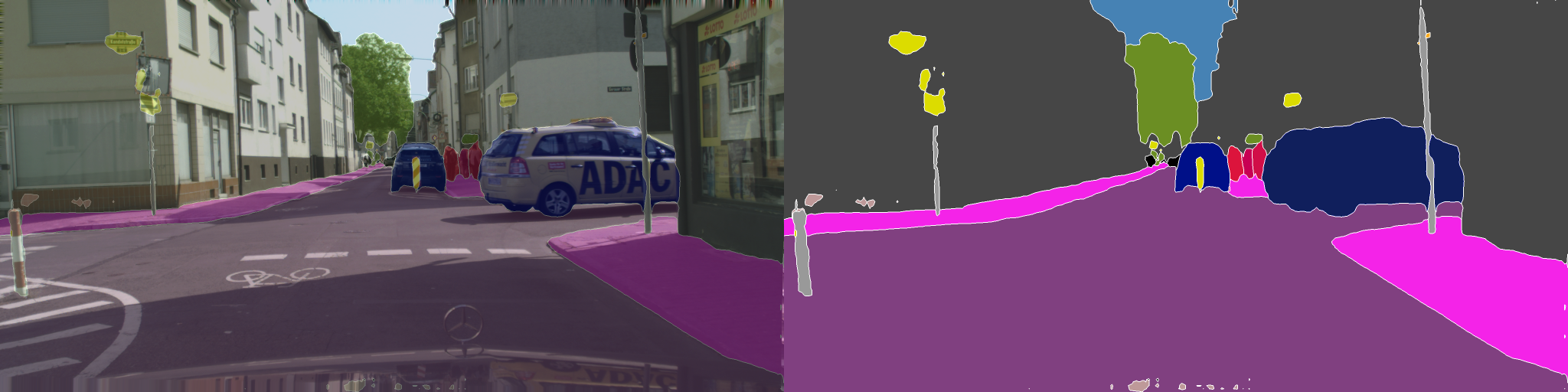}
        \includegraphics[width=1\columnwidth]{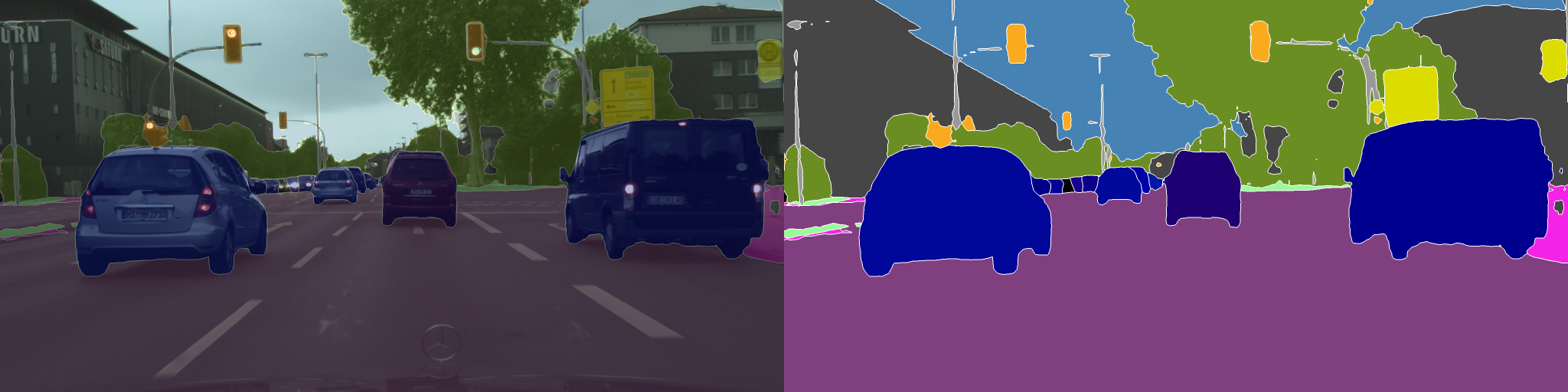}\includegraphics[width=1.0\columnwidth]{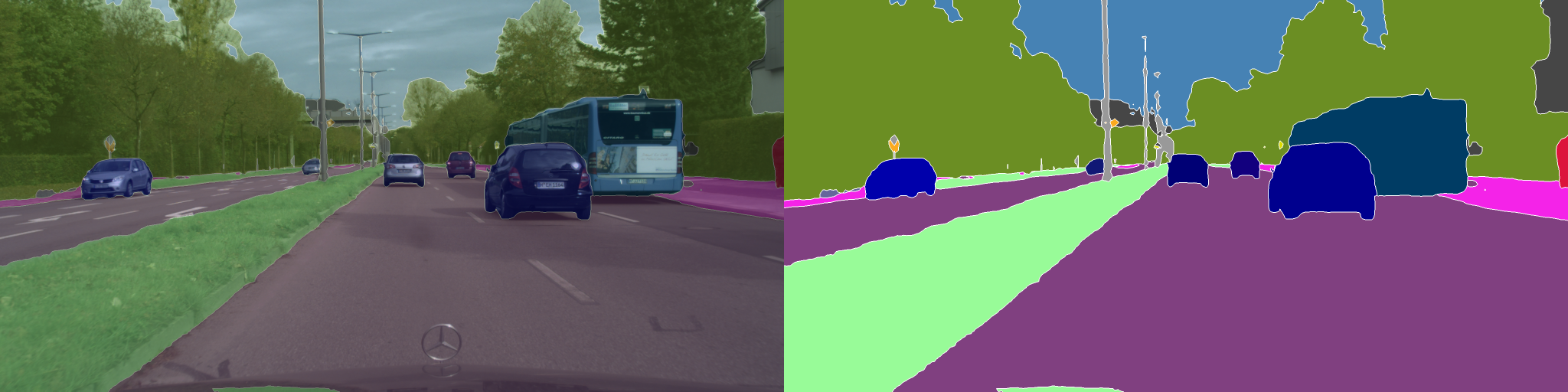}
        \includegraphics[width=1\columnwidth]{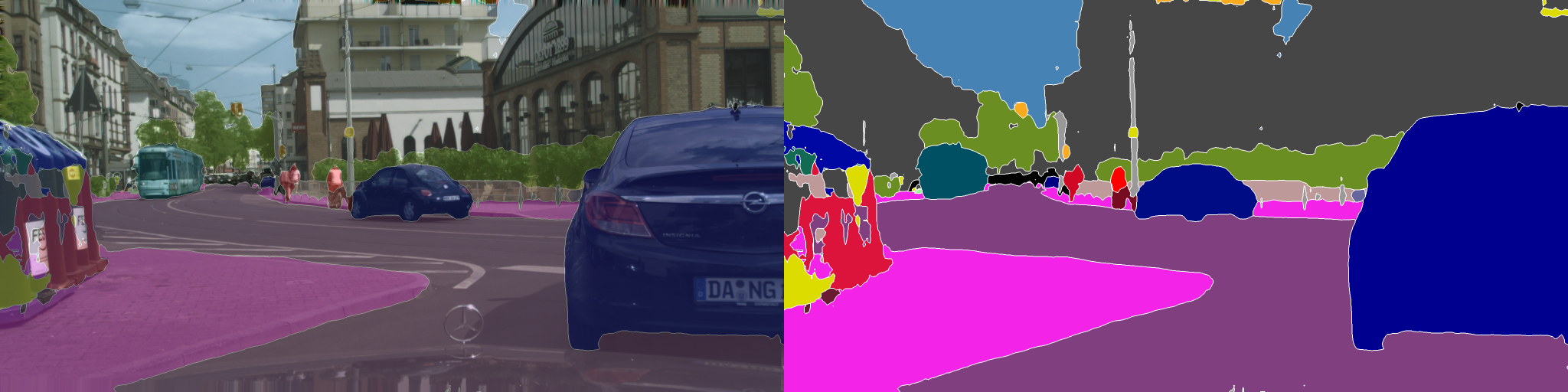}\includegraphics[width=1.0\columnwidth]{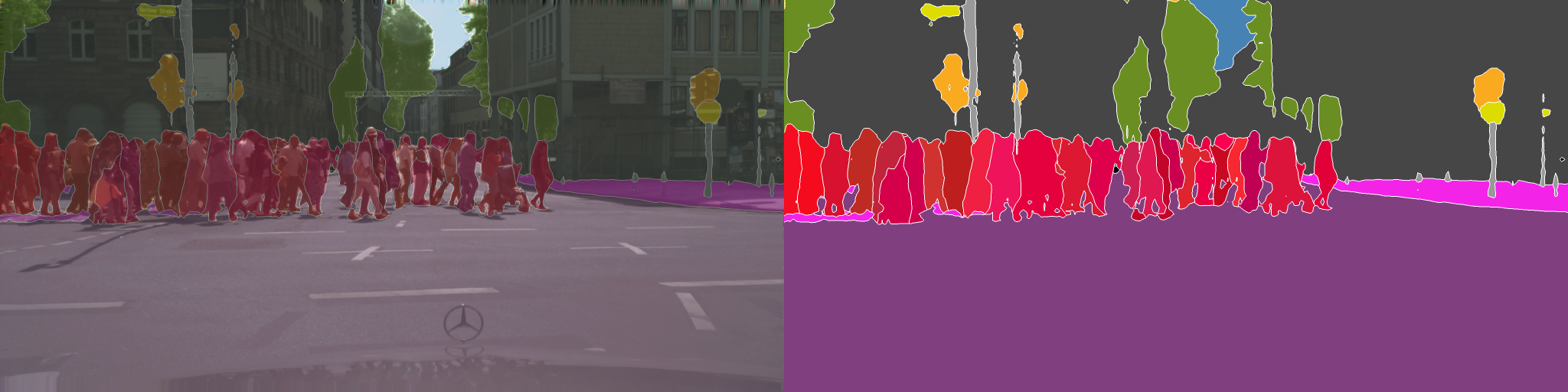}
        \caption{Visual prediction of PanoNet on Cityscapes dataset. First and third column: overlayed results; second and fourth column: corresponding panoptic segmentation results. The last row shows some failure mode.}
        \label{fig:cityscapes_results}
\end{figure*}
For semantic segmentation task, we simply adopt the same design as the original ICNet. For instance segmentation, the input and the final output layers of the ICNet framework are modified. Two more layers of normalized horizontal and vertical coordinates are appended to the RGB color channels as the input of this branch, and the final output is a 12-channel position-sensitive instance embedding map. In post-processing, a fast version of mean-shift clustering algorithm is applied class-wise to generate the instance segmentation map. More details of the PanoNet structure are discussed in the following sections.

\subsection{Loss Function}    
We design the overall loss function by dividing it into two sub-terms, one for each branch of the PanoNet. For the semantic segmentation branch, we follow the ICNet's weighted softmax cross entropy loss design. Because there are $\mathcal{T}=3$ scales of reference guidance, the loss can be defined as
\begin{align}
    \mathcal{L}_{sem} = \sum^\mathcal{T}_{t=1}\lambda_t \mathcal{L}_{sce, t}
\end{align}
$\mathcal{L}_{sce, t}$ denotes the softmax cross entropy loss of each scale, and $\lambda_t$ are the weighting factors.

For instance segmentation, we adopts the idea of discriminative loss~\cite{de2017semantic} proposed by Kirillov et al. The loss pulls the pixel embeddings closer to the mean of their cluster and pushes different clusters away from each other. We define that $C$ is the number of clusters, $N_c$, $\mu_c$ is the number of pixels and mean of  cluster $c$, $x_i$ is the embedding vector, and $\delta_v,\delta_d,\delta_r$ are the margins of the variance, distance and regularization loss terms. $[x]_{+} = \max(0, x)$ denotes the hinge function and $\lVert \cdot \rVert$ denotes the L2 distance. Then our modified instance segmentation loss can be expressed as
\begin{align}
     \mathcal{L}_{var} = \frac{1}{C}\sum_{c=1}^C\frac{1}{N_c}\sum_{i=1}^{N_c}\left[\lVert\mu_c-x_i\rVert-\delta_v\right]^2_+ \\
     \mathcal{L}_{dist} =\frac{1}{C(C-1)} \underset{_{c_a\ne c_b}}{\sum_{c_a=1}^C\sum_{c_b=1}^C}\left[2\delta_d-\lVert\mu_{c_a}-\mu_{c_b}\rVert\right]^2_+ \\
     \mathcal{L}_{reg} =\frac{1}{C}\sum_{c=1}^C\frac{1}{N_c}\left[\lVert\mu_c\rVert-\delta_r\right]^2_+ \\
     \mathcal{L}_{inst, t} = \alpha  \mathcal{L}_{var, t} + \beta  \mathcal{L}_{dist, t} + \gamma  \mathcal{L}_{reg, t} \\
     \mathcal{L}_{inst} = \sum^\mathcal{T}_{t=1}\lambda_t \mathcal{L}_{inst, t}
\end{align}

\begin{table*}[h]
    \begin{center}
    \begin{tabular}{|l|c c c| c c c | c c c |c c|c |}
    \hline
    method & PQ& SQ& RQ& PQ\textsuperscript{Th} &SQ\textsuperscript{Th} &RQ\textsuperscript{Th} &PQ\textsuperscript{St} &SQ\textsuperscript{St} &RQ\textsuperscript{St} & mIoU & AP &runtime (ms)\\
    \hline\hline
    Li et al.~\cite{Li_2018_ECCV} & 53.8 &- &- &42.5 &- &- &62.1 &- &- & 71.6&28.6&7000 \\
    MR-CNN-PSP~\cite{kirillov2018panoptic} & 58.0 & 79.2&  71.8&  52.3 & 77.9 & 66.9 & 62.2&  \textbf{80.1} & 75.4  & 75.2&32.8& 1105\\
    Panoptic FPN~\cite{kirillov2019panoptic} & 58.1  & -&  -&  52.0 & - & - & 62.5&  - & -  & \textbf{75.7}&33.0& 500 \\
    UPSNet~\cite{arxiv:upsnet}& \textbf{59.3}& \textbf{79.7} &\textbf{73.0} &\textbf{54.6}& \textbf{79.3} &\textbf{68.7}& \textbf{62.7} &\textbf{80.1}& \textbf{76.2} & 75.2&\textbf{33.3}& 236\\
    \hline
    PanoNet&  55.1&77.5 &67.9  & 46.8 &74.6 &60.9 &60.5 &79.6 & 72.9  & 74.6&23.1 & \textbf{50} \\
    \hline
    \end{tabular}
    \end{center}
    \caption{Panoptic segmentation results on Cityscapes validation set. Superscripts Th and St stand for things and stuff. Runtimes, if not stated in the original paper, are guessed in favor of the method. Metrics that are not reported are left as `-'.}
    \label{tab:panoptic_results}
\end{table*}                  

\subsection{Position-Sensitive Embedding}
Previous embedding-clustering based approach does not take the position of different instances into consideration. When there are two instances with similar appearance in a same image, such model often fails. Inspired by Liu et al.~\cite{liu2018intriguing}, we propose to add coordinate maps in addition to the origianl RGB image as the input of the network. However, achieving a weighting balance between the two aspects (position and appearance) of the final fused embedding can be tricky. Through experiments, we observe that directly training such network from scratch may cause the network to overuse the coordinate information such that the network's learned embedding does not generalize very well and performs poorly on validation set. To address this issue, we apply a two-stage training technique. More specifically, we first train the network with discriminative loss by using the pretrained semantic segmentation weights and re-initializing last few layers. After converging, the network is trained again with the two coordinate input layers added. This approach allows the network to learn how to effectively utilize the coordinates to segment different instances and fuse the position-sensitive information into the final embedding vector.
                                                                                                                                                                                                                                                
\subsection{Clustering}
In the discriminative loss function, if we choose the margins such that $\delta_d > \delta_v$, the embedding will move each vector $x_i$ closer to its own cluster center than to all other clusters. To get the instance masks, we need to choose a clustering method to group the vectors $x_i$ into their respective clusters. As the number of instances, i.e., the number of clusters in the embedding space of $x_i$'s, is unknown, the mean-shift algorithm~\cite{comaniciu2002mean} is a good fit in this case. Mean-shift is controlled by a single `bandwidth' parameter in lieu of the number of clusters. The bandwidth can be viewed as the distance threshold for determining whether a point belongs to the neighbourhood of a cluster center. Because the $\delta_v$ term in the loss function plays a similar role as the bandwidth, we choose it as the base value of the bandwidth, and we search within a small range for an optimal value of each semantic category on the training set. This is because we cannot train the discriminative loss to zero in practice, so that the actual optimal bandwidth value for clustering is slightly larger than $\delta_v$. The search process is one-time and straightforward (usually takes few hours to complete). We also accelerate the mean-shift algorithm with discretized bin-seeding (seeds are chosen from a grid of original points). The clustering process adds  minimal overhead to the whole pipeline and does not affect the real-time inference speed.

\subsection{Implementation Details}
We set the hyperparameters of the loss function as $\tau_1 = 1, \tau_2 = 0.4, \tau_3 = 0.16$ for the three layer of cascade guidance (1/4, 1/8, 1/16 scale of original image), $\delta_v = 0.25, \delta_d = 1, \delta_r = 6$ for the margins and $\alpha = 1, \beta = 1, \gamma = 0.1$ for the weights of discriminative loss. For training on Cityscapes dataset, we choose Adadelta with the learning rate of 0.003 and polynomial decay policy.

%% file: sections/4-results.tex
\section{Experiments}

We train the PanoNet on the Cityscapes train set (2975 images) and test the model on the validation set (500 images). For both training and testing, we use the original high-resolution images without down-sampling. The inference speed measurement is conducted on a single Titan X GPU.
\paragraph{Panoptic Segmentation} We report the panoptic segementation results on Cityscapes together with other state-of-art methods in Table \ref{tab:panoptic_results}. Only our method requires no object detectors or region proposals. We show that the PanoNet runs much faster than all other methods and achieves decent panoptic quality score at the same time. Our inference time is 50 ms, less than 1/4 of the fastest UPSNet. On the other hand, our PQ score is 55.1, which is even better than one method that does not pay attention to the speed. 

\begin{table*}[t]
    \begin{center}
    \begin{tabular}{|l |c  c| c c c c c c c c|}
    \hline
    semantic segm.  & AP & AP\textsubscript{0.5} & person &  rider & car & truck&  bus&  train&  mcycle & bicycle \\
    \hline\hline
    ICNet\cite{Zhao_2018_ECCV}  &23.1 &42.3 &8.8 & 8.2& 29.6&29.3 &45.3 &45.4 &9.3 &8.5 \\
    DeepLabv3+\cite{chen2018deeplab} & 19.6& 40.1& 11.9&14.2 & 28.2& 14.4 & 39.6 & 22.9 & 13.1 &12.7 \\
    \hline
ground truth & \textbf{59.7}& \textbf{77.3}&\textbf{41.8} &\textbf{73.2} &\textbf{31.8} &\textbf{78.6} &\textbf{76.8} &\textbf{71.9} & \textbf{58.0}&\textbf{45.6} \\
    \hline\hline
    \multicolumn{3}{|l|}{number of instances in train set}&17.9k &1.8k &26.9k& 0.5k& 0.4k& 0.2k& 0.7k& 3.7k \\
    \hline
    \end{tabular}
    \end{center}
    \caption{Influence of semantic segmentation results on instance segmentation performance. The last row shows the total number of training instances in the Cityscapes dataset.}
    \label{tab:sem_inst}
\end{table*}

Some visual examples are shown in Fig.~\ref{fig:cityscapes_results}. Because our method does not need to handle the conflicts of semantic and instance segmentation like other two-stage solutions, there is no large black (unlabeled) area in the images as observed in those methods. The PanoNet is also able to properly group non-connected regions, which cannot be properly segmented by some other instance clustering approaches. The last row illustrates some failure modes of our solution. The left image shows semantic segmentation error of some trash cans, and the right one shows a hard instance segmentation case where there are crowded humans with occlusion.

% As our approach can essentially utilize any semantic segmentation framework as backbone and the semantic segmentation branch is not modified, we do not report the performance here. For detailed semantic segmentation results of the ICNet, readers may refer to the original paper.
% Effect of the semantic segmentation and clustering components on the performance of our method on the Cityscapes validation set. We study this by gradually replacing each component
% with their ground truth counterpart. Row 1 vs row 3: the quality of
% the semantic segmentation has a big influence on the overall performance. Row 1 vs 2 and row 3 vs 4: the effect of the clustering
% method is less pronounced but also shows room for improvement

\begin{figure}[h]
  \centering
  \includegraphics[width=0.7\linewidth]{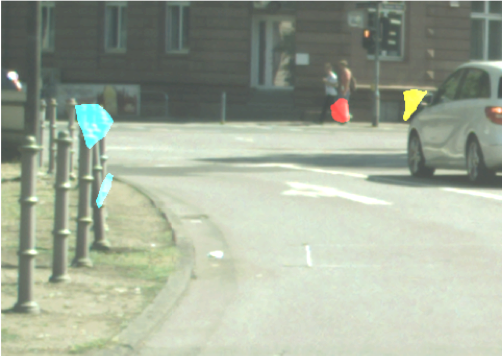}
  \caption{Zoomed-in view of mislabeled semantic regions. The highlighted regions are all mislabeled as `bicycle' class by the semantic segmentation branch. They are clustered into three `bicycle' instances, resulting in three false positives.}
  \label{fig:mislabel}
\end{figure}

\paragraph{Influence of Semantic Segmentation on Instance Segmentation}
As our method does instance clustering on semantic segmentation results, error may accumulate during this process. Compared with the UPSNet, we observe higher performance gap for instance segmentation (things) than semantic segmentation (stuff). To further understand the influence of semantic segmentation on instance segmentation, we replace the semantic segmentation map with a) prediction from a more complex model - DeepLabv3+~\cite{chen2018deeplab} and b) ground truth. These results are reported in Table \ref{tab:sem_inst}. We can see that the average precision is significantly higher when the ground truth semantic segmentation is given. This shows that our method is highly efficient on the instance clustering task. The AP is higher on bus and train class because there are usually only one or two instances of these classes in a single image (little clustering work is needed). When using semantic prediction instead of the ground truth, instance segmentation is very prone to error especially when some small regions are incorrectly mislabeled. One of such examples is shown in Fig. \ref{fig:mislabel}. The highlighted regions are mislabeled as `bicycle' class and clustered into three instances. Though these small regions won't affect the semantic segmentation metrics much, they significantly degrade the instance performance because of the introduced false positives. The `car' class has the smallest performance gap without the ground truth semantic segmentation, because this class has the most training examples in the dataset. Last, we show that, more accurate semantic segmentation methods, such as DeepLabv3+, does not necessarily lead to performance improvement. The semantic segmentation models that are conservative on `things' categories and predict fewer false positives tend to result in better instance segmentation.

\begin{table}[h!]
    \begin{center}
    \begin{tabular}{|l |l|c  c|}
    \hline
    method & backbone& AP & AP\textsubscript{GT}\\
    \hline\hline
    Discriminative  Loss\cite{de2017semantic} &ResNet38 &21.4 &37.5  \\
    PanoNet w/o pos. &ICNet& 20.5&58.9  \\
    PanoNet &ICNet& 23.1&63.7 \\
    \hline
    \end{tabular}
    \end{center}
    \caption{Performance of instance segmentation. AP\textsubscript{GT} means the average precision of instance segmentation given the ground truth semantic segmentation. }
    \label{tab:position_embedding}
\end{table}

\paragraph{Position-Sensitive Embedding} To study whether adding position-sensitive component to the embedding leads to any improvement on instance segmentation task, we train a variant of PanoNet under the same settings but without providing position information (i.e., removing coordinate map from input). The results are reported in Table \ref{tab:position_embedding}. We observe that position-sensitive embedding increases the average precision. Even with a much more light-weight backbone (ICNet compared to ResNet38), PanoNet outperforms the baseline model.

\begin{figure}[!h]
  \centering
  \includegraphics[width=0.48\linewidth]{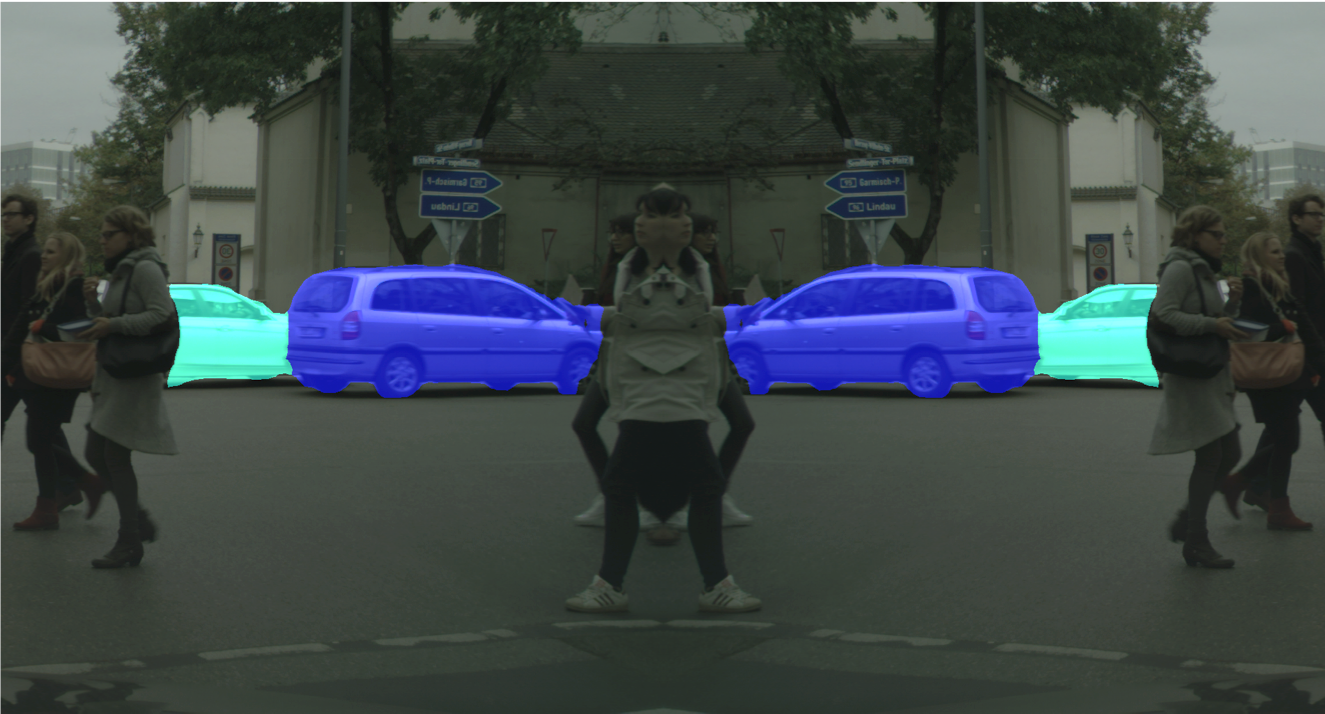}
  \includegraphics[width=0.48\linewidth]{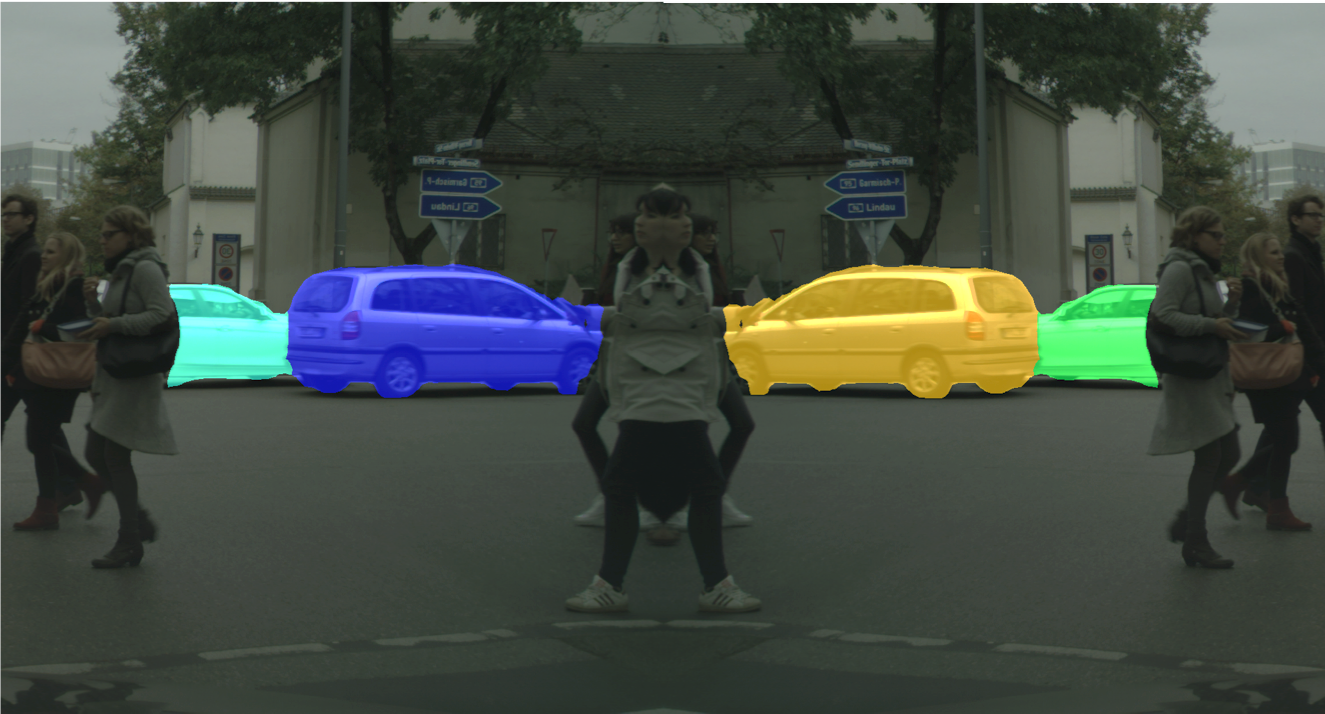}
  \caption{Instance segmentation on an artificially created image (mirrored about the axis in the middle). Left: prediction result of non-position-sensitive embedding, the two mirrored cars are grouped into one single instance; right: prediction result of position-sensitive embedding.}
  \label{fig:pos_reflected}
\end{figure}

 Position information is especially useful in scenarios where two instances have similar or the same appearance. Fig.~\ref{fig:pos_reflected} shows such an example. The image was created by mirroring the left half of the original image to the right half. If we use pure feature embedding to cluster instances, the two mirrored cars cannot be distinguished from each other as expected. But when position information is provided, all the four instances can be correctly predicted as shown in the figure.

\paragraph{Runtime and Memory Analysis} We report the computational complexity of our model in Table \ref{tab:speed_and_mem}. PanoNet shows excellent performance-accuracy trade-off. To our best knowledge, PanoNet is the first panoptic segmentation model that runs in real time on a single GPU with 2-mega-pixel high-resolution input images. 

Our model has advantages over other methods for three main reasons. First, for applications such as autonomous driving and augmented reality, real-time performance is highly desired. The inference speed of PanoNet is at 20 fps and significantly faster than all other approaches, making it favorable especially when hardware resources are limited. Second, the highest-end GPUs on the market usually have 12 GB memory, while many other consumer GPU products have only 4 GB to 8 GB memory. For high-resolution images, the listed models usually have to crop the image into several smaller sub-images to process separately simply because the model cannot be fit into the GPU's memory. This makes PanoNet stand out as it only requires 3GB memory for images as large as $2048\times1024$. Thus, our whole model can be easily fit into most GPUs so that the output of a complete image is generated in a single shot. Last, PanoNet is easy to train beacause of it contains many fewer parameters. Most other panoptic segmentaion models use 16 GPUs or more \cite{arxiv:upsnet} to train, while our training process only takes 4 GPUs thanks to the compactness of the model.

\begin{table}[]
    \begin{center}
    \begin{tabular}{|l |c|c|c|  c|}
    \hline
    method &PQ& \# p.& mem. & fps\\
    \hline\hline
    Li et al.~\cite{Li_2018_ECCV}&53.8& -&48 GB & $<0.5$\\
    MR-CNN-PSP~\cite{kirillov2018panoptic} &58.0&92M & $>12$ GB &0.9  \\
    Panoptic FPN~\cite{kirillov2019panoptic} &58.1& - & $>12$ GB &2  \\
    UPSNet~\cite{arxiv:upsnet}&59.3&46M& -&4.2  \\
    DeeperLab (Xception-71)~\cite{yang2019deeperlab}& 56.5 & - & -&3.2  \\
    DeeperLab (Wider MNV2)~\cite{yang2019deeperlab}&52.3&-& -&6.7  \\
    DeeperLab (Light Wider MNV2)~\cite{yang2019deeperlab}&48.1&-& -&10.2  \\
    DeeperLab (Light Wider MNV2)~\cite{yang2019deeperlab}&39.3&-& -&24  \\
    PanoNet &55.1&12M& 3 GB&20 \\
    \hline
    \end{tabular}
    \end{center}
    \caption{Panoptic quality (PQ) and computational complexity comparison of PanoNet and other methods on full-scale Cityscapes images. `\# p.' means the number of parameters of the model. Metrics that are not reported are left as `-'.}
    \label{tab:speed_and_mem}
\end{table}

\paragraph{Sharable Weights} Currently our PanoNet model consists of two independent ICNet branches. However, it is possible for these two branches to share the weights of shallow layers to further compress the model. We train a variant of PanoNet with shared weights except for the final three layers leading to the prediction. This yields 52.3 PQ, which is 5\% lower than the baseline. As PanoNet is already light-weight and fast, further study on weight sharing and model compression is beyond the scope of this paper.

% \begin{table}[h]
%     \begin{center}
%     \begin{tabular}{|l |c  c|c|}
%     \hline
%     method & AP & AP\textsubscript{0.5} & fps\\
%     \hline\hline
%     Discriminative  Loss &17.5 &35.9 & 5 \\
%     Mask-R-CNN & 32.0&58.1 & $<1$  \\
%     Box2Pix & 13.1&27.2 & 11 \\
%     PanoNet & 13.1&27.2 & 20 \\
%     \hline
%     \end{tabular}
%     \end{center}
%     \caption{Results.   Ours is better.}
% \end{table}

%% file: sections/5-conclusion.tex
\section{Conclusion}

In this paper, we have proposed a real-time panoptic segmentation method PanoNet. It utilizes same network structure for both semantic and instance segmentation. Because of the simplicity of its framework and  effectiveness of position-sensitive embedding, PanoNet significantly increases inference speed without much performance sacrifice. Our work provides a practical solution to many real-world tasks that require fast scene segmentation on high-resolution images. In the future, we would like to explore better training strategies and framework design to increase the robustness of instance segmentation to semantic segmentation error.